\documentclass[lettersize,journal]{IEEEtran}
\usepackage{amsmath,amsfonts}
\usepackage{algorithmic}
\usepackage{algorithm}
\usepackage{array}
\usepackage[caption=false,font=normalsize,labelfont=sf,textfont=sf]{subfig}
\usepackage{textcomp}
\usepackage{stfloats}
\usepackage{url}
\usepackage{verbatim}
\usepackage{graphicx}
\usepackage{cite}
\usepackage{color}
\usepackage[svgnames]{xcolor}  
\usepackage{amssymb}  
\usepackage{multirow}  
\usepackage{threeparttable}  
\usepackage{tabularx}
\usepackage{longtable}
\usepackage[colorlinks,linkcolor=blue,citecolor=blue]{hyperref}  

\usepackage[table]{xcolor}
\usepackage{orcidlink}  

\usepackage{soul}  

\begin{document}

%
\title{Pushing the Limit of Finetuning in Active Finetuning: \protect\\A Simple yet Effective Three-Phase Framework with Hierarchical Gradient Harmonization}
%
%
\title{Pushing the Limit of Finetuning in Active Finetuning: \protect\\A Simple yet Effective Three-Phase Framework with Hierarchical Parameter Adaptation}
\title{Rethinking Finetuning in Active Finetuning: FACT \protect\\-- A Three-Phase Hierarchical Framework for Efficient and Robust Vision Adaptation}
\title{FACT: Rethinking Parameter Adaptation in Active Finetuning \protect\\-- A Hierarchical Three-Phase Framework for Efficient and Robust Vision Models}
%
%
\title{FACT: Rethinking Active Finetuning with an Efficient and Robust Hierarchical Framework}
\title{FACT: A Simple and Efficient Framework for Active Finetuning}

\author{Wenshuai Xu\orcidlink{0009-0001-1336-9929}, You Song\orcidlink{0000-0003-0236-859X}, Yuzhuo Cui\orcidlink{0009-0002-3857-5486}, Minjie Ren\orcidlink{0000-0002-6080-8074}, Qingjie Liu\orcidlink{0000-0002-5181-6451},~\IEEEmembership{Member,~IEEE}, and Zhenghui Hu\orcidlink{0000-0002-6106-0416}
%
%
\thanks{This research was supported in part by ``Pioneer'' and ``Leading Goose'' R\&D Program of Zhejiang (No. 2024C01020), in part by the National Natural Science Foundation of China (No. 62302031), and in part by the Zhejiang Provincial Natural Science Foundation of China (Nos. LQ23F020024 and LZJMZ24D050009).}
\thanks{Corresponding author: Qingjie Liu. (e-mail: qingjie.liu@buaa.edu.cn)}}


\markboth{ IEEE TRANSACTIONS ON IMAGE PROCESSING,~Vol.~00, No.~0, May~2025}%
{Shell \MakeLowercase{\textit{et al.}}: A Sample Article Using IEEEtran.cls for IEEE Journals}

\IEEEpubid{0000--0000/00\$00.00~\copyright~2025 IEEE}

\maketitle

\begin{abstract}
The main goal of active finetuning is to improve a pretrained model's performance on a specific task or domain by finetuning it with carefully selected informative or challenging data. Previous research has predominantly focused on the \textit{active} aspect (i.e., data selection) while uniformly employing \textit{full finetuning} for model adaptation, which inevitably distorts pretrained features due to distribution shift. This issue becomes particularly pronounced when the model size is large relative to the finetuning data quantity, leading to heightened overfitting risks. To address this critical gap, we formally outline the \textit{\textbf{FiAF}} task that emphasizes systematic exploration of finetuning methodologies in active learning. We propose \textbf{FACT}, a three-phase hierarchical finetuning framework featuring both efficiency and simplicity, specifically designed for active finetuning scenarios. Our comprehensive experiments span: (1) Three major dataset categories encompassing classic (CIFAR10, CIFAR100, ImageNet-1k), imbalanced (CIFAR10-LT, CIFAR100-LT), and fine-grained (StanfordCars, FGVCAircraft) image classification datasets, each evaluated under 3-5 distinct sampling ratios; (2) Diverse pretrained architectures including Convolutional Neural Network (ConvNeXt), Vision Transformer (ViT), and Vision LSTM (ViL) networks; (3) A systematic investigation of frozen feature augmentation (FroFA) strategies. (4) A comprehensive and rigorous analysis of efficiency and generalizability. The results demonstrate significant improvements with strong generalization and robustness. Notably, under low sampling ratios, our framework achieves remarkable performance gains of over 20\% on the ViT model for CIFAR10, CIFAR100, and ImageNet-1k benchmarks. This systematic approach establishes new state-of-the-art performance while maintaining parameter efficiency, proving particularly effective when labeled data is scarce.
\end{abstract}


\section{Introduction}
\label{sec:intro}
\IEEEPARstart{D}{eep} learning has made significant progress in the field of computer vision that is typically attributed to the use of large-scale models and datasets. However, training such models from scratch is a time-consuming process and demands extensive amount of data. To address this, the pretraining-finetuning paradigm~\cite{TCEN_AAAI20,code_AAAI22,Equi-tuning_AAAI23} has been recognized as a favorable approach for both vision and language tasks. For vision tasks, a model can be first trained on abundant supervised or unsupervised data and be saved as a pretrained vision model (PVM). Then, the PVM is finetuned on a labeled dataset for a specific downstream task. By capitalizing on ample pretraining data and conserving valuable training resources during the finetuning phase, this paradigm has archived remarkable adoption in practical applications.

\begin{figure}[t]
\centering
\includegraphics[width=1.0\columnwidth]{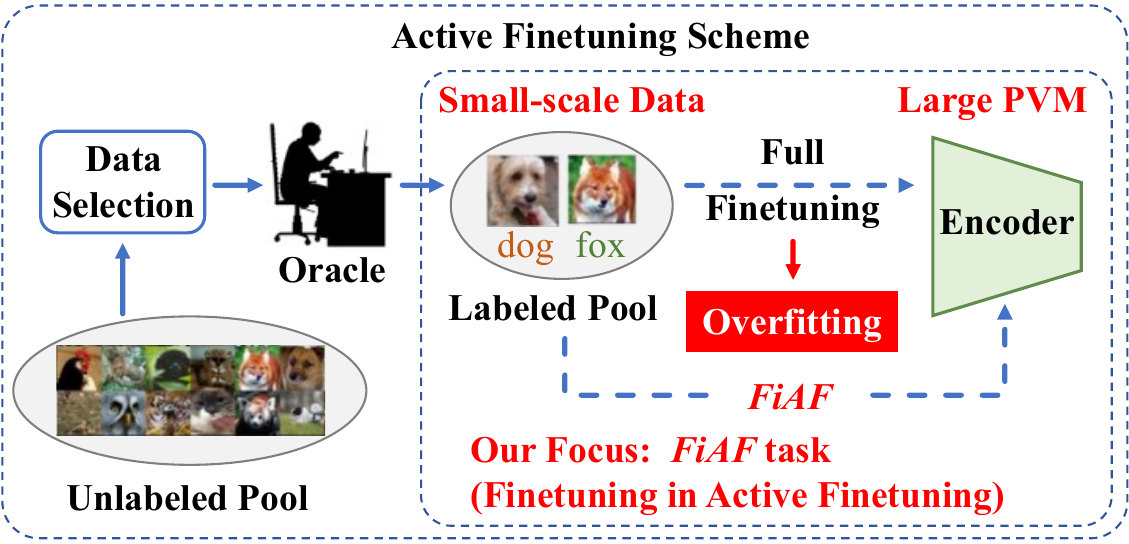}
\caption{We focus on the \textbf{F}inetuning task \textbf{i}n \textbf{A}ctive \textbf{F}inetuning (\textit{\textbf{FiAF}}). Prior studies have emphasized the \textit{active} (i.e., data selection) over the \textit{finetuning}. However, the full finetuning in \textit{FiAF} faces overfitting challenges due to small-scale data and large pretrained vision model (PVM). The compatibility of finetuning methods in the active finetuning scheme has often been overlooked in previous research.}
\label{fig:how_to_finetune}
\end{figure}

Although the pretraining-finetuning paradigm has significantly improved training efficiency and downstream task performance, researchers continue to seek ways to optimize the use of a limited annotation budget by selecting the most informative samples from a large pool of unlabeled data~\cite{liu2023understanding}.
Active learning algorithms are promising for optimizing limited annotation budgets~\cite{LLoss_CVPR2019,BADGE_ICLR2020,TA-VAAL_CVPR2021,ALFA-Mix_CVPR2022,noiseStability_AAAI2024} but struggle in the pretraining-finetuning paradigm~\cite{SSmeetAL_ICCV2021,ActiveFT_CVPR2023}. The iterative batch selection strategy, which begins with a random subset and sequentially adds samples until the budget is exhausted, introduces biases when applied with small batch sizes. These biases can lead to overfitting and degrade the quality of the pretrained model's representations, especially given the typically constrained annotation budgets in this context.

\IEEEpubidadjcol
%
Active finetuning methods~\cite{ActiveFT_CVPR2023,xu2024activedc,BiLAF_NIPS24,VeCAF_MM24} address the shortcomings of traditional multi-round iterative active learning in the pretraining-finetuning paradigm by selecting a subset that fits the entire distribution at once. However, these methods primarily focus on data selection and often use the currently popular ViT series models for full parameter finetuning during the finetuning phase, overlooking the compatibility of the finetuning method with the active finetuning task. This lack of compatibility primarily arises because the number of training samples selected through active finetuning is typically small, and their distribution shift adversely affects the pretrained model, especially when the pretrained model has strong representation ability~\cite{LP-FT_ICLR2022,FT4FSL_ICLR2023,mix_cd}. Moreover, when the sample size is small but the model parameters are numerous, the training efficiency decreases, and the risk of overfitting increases~\cite{HDSSL_ijcai2017,pmf_CVPR2022}.

%
Prior studies have not adequately addressed the impact of the data selection on the subsequent finetuning phase, focusing more on the \textit{active} than on the \textit{finetuning}.
To fill in this gap, we refer to the task of \textbf{F}inetuning \textbf{i}n \textbf{A}ctive \textbf{F}inetuning as \textit{\textbf{FiAF}}. 
%
We analyze previous finetuning methods in the \textit{FiAF} task. The most common method is full finetuning, however, this approach underperforms due to distorted finetuning features~\cite{LP-FT_ICLR2022}. 
This distortion occurs because a randomly initialized linear classifier changes the extracted in-distribution (ID) and out-of-distribution (OOD) features inconsistently. This inconsistency arises when the classifier attempts to fit the finetuning samples.
Kumar et al. address this distortion problem by using a linear classifier that is well-trained on finetuning samples~\cite{LP-FT_ICLR2022}. Although this method is effective, it is primarily designed for large-scale finetuning datasets, which contain sufficient samples to provide an unbiased estimation. In contrast, in the \textit{FiAF} tasks, the limited number of finetuning samples and overparameterized model lead to a biased estimation of the updated parameters.

To address the incompatibility of existing finetuning methods in \textit{FiAF}, we propose \textbf{FACT} (\textbf{F}inetuning in \textbf{ACT}ive finetuning), a targeted three-phase hierarchical finetuning framework.
Our finetuning framework is structured into three sequential phases. In the first phase, the pretrained feature extraction model is frozen, and only the task-specific head is finetuned (linear probing, LP). In the second phase, both the pretrained feature extraction model and the task head finetuned in the first phase are jointly finetuned (full finetuning, FF). In the third phase, the feature extraction model finetuned in the second phase is used to extract features from the finetuning data, which are then trained and classified using lightweight model combined with the frozen feature augmentation (FroFA) algorithm.
%
The first and second steps employ data-friendly finetuning strategies tailored for ID and OOD scenarios. This increased tolerance to selected and unselected samples enhances the robustness of active finetuning. The third step addresses the overfitting risk arising from a limited finetuning dataset and a large number of model parameters.
Extensive experiments have validated the effectiveness and efficiency of our framework.
%
%

Our contributions are summarized as follows:

\begin{itemize}
\item We observed that finetuning in active finetuning differs from traditional finetuning due to the distribution shift and limited size of the finetuning dataset. To address the problems arising from this discrepancy, we outline a more focused task, \textit{FiAF}, aimed at mitigating these issues.
\item We propose FACT, a three-phase hierarchical finetuning framework that addresses the issues with existing finetuning methods in the \textit{FiAF} task through FF following LP, frozen feature augmentation algorithm, as well as the use of well-regularized lightweight models. This is a more optimal approach for the \textit{FiAF} task.
\item We expanded our evaluation of various backbone networks for the active finetuning task and found that the Vision LSTM architecture demonstrated strong performance in this context.
\item We conducted experiments on well-established classification datasets, long-tail datasets, and fine-grained datasets. Our approach not only achieves superior performance but also demonstrates higher training efficiency.
\end{itemize}

\begin{figure*}[t]
\centering
\includegraphics[width=1.00\textwidth]{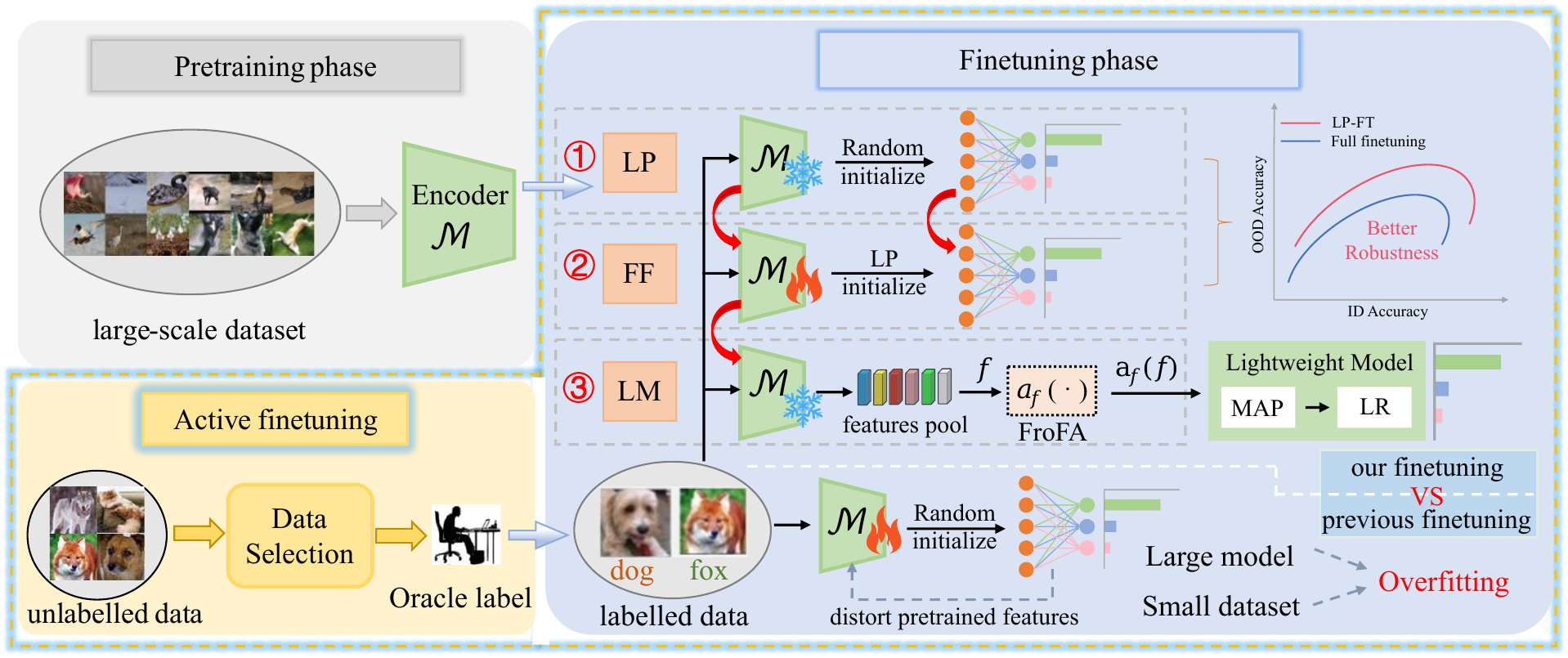} 
\caption{
The proposed \textbf{FACT} framework is specifically designed for the \textit{FiAF} task to address the problems of data distribution shift and over-parameterization of pretrained models during the finetuning phase. We highlight it with a light blue background in the figure, involves a three-phase process: 
\textbf{(1) Linear Probing (LP):} In this phase, the parameters of the pretrained model are frozen. Only the linear classifier, initialized with random parameters, is finetuned. 
\textbf{(2) Full Finetuning (FF):} The linear classifier finetuned in the first phase is integrated into the pretrained model, and all parameters are finetuned jointly. 
\textbf{(3) Lightweight Model (LM):} The model finetuned in the second phase is used to extract features from the finetuning data. These features, processed through frozen feature augmentation (FroFA), serve as input for training a lightweight model. 
The lightweight model consists of a multi-head attention pooling (MAP) and a linear layer with logistic regression (LR) algorithm.
Compared to previous methods, our approach has better robustness and generalization.
}
\label{fig:THriFT}
\end{figure*}

\section{Related Work}
\label{sec:related_work}

\subsection{Active Finetuning}
Traditional active learning algorithms are primarily designed for training from scratch, and their batch iteration method with small sample sizes faces challenges in the pretraining-finetuning paradigm~\cite{SSmeetAL_ICCV2021,TypiClust_ICML2022,ActiveFT_CVPR2023}.
In response, ActiveFT~\cite{ActiveFT_CVPR2023} has been specifically developed for this context. It selects data by aligning the distribution of the selected samples with that of the original unlabeled pool in the feature space.
Building on this foundation, ActiveDC~\cite{xu2024activedc} employs a novel data augmentation algorithm to calibrate the distribution of selected samples by leveraging implicit category information from the unlabeled data pool.
%
Lu et al. proposed the BiLAF~\cite{BiLAF_NIPS24} algorithm, which enhances active finetuning methods by integrating diversity and uncertainty into the selection process. This is achieved by combining the center sample selection method with the uncertainty boundary sample selection method.
Zhang et al. proposed Vision-language Collaborative Active Finetuning (VeCAF) method~\cite{VeCAF_MM24}, which leverages the semantically rich language embedding space of text encoders, such as CLIP~\cite{CLIP_ICML2021} and BLIP-2~\cite{Blip2_ICML2023}, to extract text embeddings from image captions. 
VeCAF combines text embeddings with a cross-attention embedding enhancement to refine the image features extracted by PVM. This refinement enables the enhanced features to better capture the rich semantic information of the training samples.

\subsection{Finetuning  Architectures} 
Rather than training a model from scratch, a pretrained model is used as a starting point for further finetuning on a specific task or dataset.
By capturing general patterns and features from a large dataset, the pretrained model serves as a knowledge base, providing a robust foundation for efficiently addressing new, similar problems.
In research related to active finetuning, the model architecture typically employs ViT~\cite{ViT_ICLR2021} series models, such as DeiT~\cite{DeiT_ICML2021}, during both the pretraining and finetuning phases. 
%
While these models are well-suited for large-scale pretraining, they suffer from over-parameterization when finetuning on smaller datasets selected by active finetuning methods.
Although ViT models do not train well on the smaller finetuning datasets compared to smaller architectures, they generally excel when trained on large pretraining datasets~\cite{pmf_CVPR2022}.
In addition to the ViT architecture, excellent CNN and advanced RNN models, such as ConvNeXt~\cite{ConvNeXt_CVPR2022} and Vision LSTM~\cite{ViL_ICLR25}, are worth exploring in this field.

\subsection{Finetuning  Strategies}
Chen et al. introduced Baseline++~\cite{BaselinePP_ICLR2019}, demonstrating that a retrained linear classifier could achieve performance comparable to meta-learning methods when using a pretrained feature extractor. Similarly, S2M2~\cite{S2M2_WACV2020} demonstrated that pretraining feature extractors with techniques like Rotation Augmentation, Manifold Mixup, and Knowledge Distillation enhances robustness~\cite{AutoAug, TrivialAug}. These Linear Probing methods involve retraining only the linear classifier while keeping the pretrained feature extractor fixed, based on the assumption of strong robustness~\cite{P-Transfer_AAAI2021}. However, this assumption may not always hold. 
Kumar et al. suggest that the easy two-step strategy of linear probing then full finetuning, LP-FT~\cite{LP-FT_ICLR2022}, sometimes used as a finetuning heuristic, combines the benefits of both finetuning and linear probing. Empirically, LP-FT outperforms both finetuning and linear probing on many datasets. LP-FT can mitigate tradeoffs between ID and OOD accuracy in their context.
Beyond these, parameter‑efficient finetuning (PEFT) methods have gained popularity~\cite{LoRA_plus_ICML,LoRA-GA_2024,Q-LoRA_2024}. Low-Rank Adaptation (LoRA)~\cite{LoRA_ICLR2022} efficiently adapts large models by applying low-rank decomposition to their weight matrices. 
Other examples~\cite{survey_peft} include prompt tuning \cite{VPT_ECCV2022}, which guides model outputs by adding learnable prompt vectors; prefix tuning \cite{PrefixTuning_ACL2021}, which adjusts model behavior by inserting trainable prefix vectors into each layer of the model; and PointGST~\cite{PointGST_TPAMI25}, which use spectral-domain structured transformations.
In contrast to these PEFT methods, our proposed FACT is a three‑phase hierarchical framework designed for FiAF under extreme low budgets and distribution shift, prioritizing adaptation and overfitting mitigation over parameter efficiency while remaining compatible with PEFT techniques as plug‑in components.
%
%

\section{Methodology}
\label{sec:method}

%
%
This section introduces our novel finetuning framework for active finetuning. Section~\ref{sec:method:setup} defines the finetuning task within active finetuning~(\textit{FiAF}), and Section~\ref{sec:method:overview} provides an overview of our approach. Section~\ref{sec:method:full_FT} then offers a theoretical analysis of the limitations of traditional full finetuning methods for \textit{FiAF} tasks. Finally, Section~\ref{sec:method:THriFT} describes the implementation of our approach and explains the role of each component.

\subsection{Problem Formulation}
\label{sec:method:setup}

Given a pool of unlabeled data $\mathcal{U}$ and an initially empty labeled pool $\mathcal{L}$, active finetuning aims to select a portion of data in one shot from $\mathcal{U}$, based on an annotation budget $\mathcal{B}$, within the pretraining-finetuning paradigm. The selected data $\{X_N\}$ is annotated by oracles and added to the labeled pool: $\mathcal{L} \leftarrow \mathcal{L} + \{ X_N, Y_N \}$, where $\{Y_N\}$ is the new annotation for $\{X_N\}$. The updated $\mathcal{L}$ is then used to train a task model composed of a feature extractor $f$ followed by a task head $g$, parameterized by $\theta$ and $\Phi$, respectively. We use $f(.; \theta)$ and $g(.; \Phi)$ to denote their feed-forward operations. Since active finetuning operates within the pretraining-finetuning paradigm, the parameters $\theta$ of the feature extractor are pretrained on a large-scale dataset, while the parameters $\Phi$ of the task head are randomly initialized for the new downstream task. Note that the feature extractor $f$ is typically a deep neural network with high-dimensional parameters $\theta \in \mathcal{R}^n$. In contrast, the task head $g$ usually has a shallow architecture, such as a single fully connected layer for classification or regression problems. Finetuning in active finetuning (\textit{FiAF}) is defined as $ \hat{\theta}, \hat{\Phi} \leftarrow f \cdot g(\theta, \Phi, \mathcal{L}) $. Here, $\hat{\theta}$ and $\hat{\Phi}$ are the updated parameters corresponding to $\theta$ and $\Phi$.

\subsection {Overview}
\label{sec:method:overview}
The overview of our proposed \textbf{FACT} framework for the \textit{FiAF} task is shown in Figure~\ref{fig:THriFT}, involves a three-phase process: 
(1) Linear Probing: In this phase, the parameters of the pretrained model are frozen. Only the linear classifier, initialized with random parameters, is finetuned. 
(2) Full Finetuning: The linear classifier finetuned in the first phase is integrated into the pretrained model, and all parameters are finetuned jointly. 
(3) Lightweight Model: The model finetuned in the second phase is used to extract features from the finetuning data. These features, processed through frozen feature augmentation (FroFA), serve as input for training a lightweight model. 
This structured framework refines the model's performance step-by-step.
The first and second steps employ data-friendly finetuning strategies tailored for both selected and unselected samples. The third step mitigates the risk of overfitting -- stemming from a limited finetuning dataset and an overparameterized model -- by incorporating a well-regularized lightweight model with FroFA. This structured finetuning process enables our approach to achieve both efficiency and effectiveness.

\subsection{Full Finetuning for \textit{FiAF}}
\label{sec:method:full_FT}
In the \textit{FiAF} task, current finetuning methods employ full finetuning. That is, the parameters $\theta$ and $\Phi$ mentioned in the ``Problem formulation'' section are updated simultaneously during finetuning. However, given that the labeled data selected by active finetuning is typically limited (i.e., $\mathcal{B}$ is relatively small), full finetuning is not ideal for the \textit{FiAF} task. This suboptimality will be demonstrated both theoretically and empirically below.

\textit{FF achieves worse OOD accuracy than LP:} 
We define full finetuning as FF and linear probing as LP. FF updates all model parameters on the new data task, while LP updates only the last linear layer (task head). That is, LP freezes the pretrained model parameters $\theta$ and updates only the task head parameters $\Phi$.
Kumar et al. claim that FF leads to better in-distribution (ID) accuracy but can achieve worse out-of-distribution (OOD) accuracy than LP when the pretrained features are strong and the distribution shift is large~\cite{LP-FT_ICLR2022}.

FF distorts pretrained features. That is, representations change exclusively within the ID subspace (i.e., subspace spanned by the training data), while remaining invariant in the orthogonal subspace. 
To see this, we take the derivative of the training loss $ L(g(f(\mathbf{X}, \theta), \Phi), \mathbf{Y}) = {\left \| \mathbf{X}\theta^T\Phi - \mathbf{Y} \right \| }^2_2 $. The gradients of the training loss with respect to the parameter $\theta$ of the feature extractor $f$ is computed as:
\begin{equation}
\label{eq:FF_distort_features}
    \nabla_\theta L(g(f(\mathbf{X}, \theta), \Phi), \mathbf{Y}) = 2\Phi(\mathbf{Y} - \mathbf{X}\theta^T \Phi)^T \mathbf{X}
\end{equation}
Let $\mathbf{S}$ denote the subspace spanned by the training samples $\mathbf{X}$, that is, $\mathbf{S} = rowspace(\mathbf{X})$. $f$ is assumed to be a linear model; $\mathbf{Y}$ denotes the one-hot labels of $\mathbf{X}$.
With Eq.~(\ref{eq:FF_distort_features}), if $u$ is a direction orthogonal to the training subspace $\mathbf{S}$, then $\nabla_\theta L(g(f(\mathbf{X}, \theta), \Phi), \mathbf{Y})u = 2\Phi(\mathbf{Y} - \mathbf{X}\theta^T \Phi)^T (\mathbf{X} \cdot u) = 0$, so the gradient updates to $\theta$ do not modify $\theta_u$ for $u \in \mathbf{S}^\perp $. However, the gradient is non-zero for directions $u$ in the ID subspace and the corresponding features $\theta_u$ change across the finetuning process. This leads to the distorted features extracted by the finetuned $f(\cdot, \theta)$ because the ID and OOD features inconsistently change.
During FF, pretrained features are only scaled (not distorted) iff $f_0$ exactly aligns with the ID subspace.

Intuitively, if the pretrained features are good, LP learns a near optimal linear head which has small OOD error but FF has high OOD error~\cite{FLYP_CVPR2023}.
The samples selected for finetuning the pretrained feature extractor are often OOD, especially in active finetuning tasks with a small annotation budget. 
Therefore, in most cases, LP is a better finetuning method than FF in active finetuning task.
%
We acknowledge that this linear analysis is a simplification; empirical validation on deep nonlinear models is provided in Table V and Section IV-D.

\subsection{Our FACT for \textit{FiAF}}
\label{sec:method:THriFT}


\subsubsection{LP-FT balances ID and OOD data finetuning }
Kumar et al. theoretically demonstrate that the tradeoff between ID and OOD accuracy arises even in a simple setting: finetuning overparameterized two-layer linear networks. Their analysis suggests that the straightforward two-step strategy of linear probing followed by full finetuning, LP-FT~\cite{LP-FT_ICLR2022}, often used as a finetuning heuristic, combines the benefits of both LP and FF.
\begin{equation}
\label{eq:lpft}
\begin{aligned} & \forall t,L_{\mathrm{ood}}(\theta_{\mathrm{ft}}(t)^{\top}\Phi_{\mathrm{ft}}(t))>0,if~\Phi_0\sim\mathcal{N}(0,\sigma^2I),\\  & \forall t,L_{\mathrm{ood}}(\theta_{\mathrm{ft}}(t)^{\top}\Phi_{\mathrm{ft}}(t))=0,if~\Phi_0=\Phi_{\mathrm{lp}}^{\infty}.
\end{aligned}
\end{equation}
Under non-degeneracy conditions, as shown in Eq.~\eqref{eq:lpft}, LP-FT theoretically mitigates OOD errors more effectively than FF. This makes LP-FT preferable for active finetuning tasks aimed at improving prediction accuracy on unseen, including unselected, samples by finetuning selected ones.

\subsubsection{Lightweight Model with FroFA}

We use a lightweight model following frozen feature augmentation (FroFA)~\cite{FroFA_CVPR2024} to address challenges associated with small-scale datasets and model over-parameterization.
%
Our experiments indicate that training a linear classifier or lightweight model on top of PVM outputs, a.k.a.~frozen features, leads to impressive performance on active finetuning tasks.
This discovery is consistent with some understanding in previous literature~\cite{BiT_ECCV2020, CLIP_ICML2021, ScalingViT_CVPR2022}.
Kolesnikov et al. and Zhai et al.  reveal that training a lightweight model on frozen features from a large-scale pretrained backbone yields high performance across various downstream tasks~\cite{BiT_ECCV2020, ScalingViT_CVPR2022}.

With the recent success of frozen feature augmentation (FroFA) techniques in few-shot tasks and the growing need for lightweight models to effectively train data features, embedding FroFA into our finetuning task has become a straightforward and efficient solution.
%
%
The FroFA processing flow consists of the following key steps:

\begin{itemize}
    \item Feature Reshaping: The initial 2D feature \( f \in \mathbb{R}^{N \times C} \) is reshaped into a 3D form \( f^* \in \mathbb{R}^{\sqrt{N} \times \sqrt{N} \times C} \), where \( N \) is the number of tokens and \( C \) is the number of channels.
    \item Channel Separation: Each channel \( c \) is treated as a separate 2D spatial representation \( f^*_c \in \mathbb{R}^{\sqrt{N} \times \sqrt{N} \times 1} \).
    \item Value Normalization (\(t_{f \to x}\)): The reshaped features are normalized into an image-like value range (e.g., \([0,1]\)) using the transformation \( x_f = (f^* - f_{\text{min}})/(f_{\text{max}} - f_{\text{min}}) \).
    \item Inversion Mapping (\(t_{f \leftarrow x}\)): An inverse mapping \( f^* = x_f \cdot (f_{\text{max}} - f_{\text{min}}) + f_{\text{min}} \) restores the features to their original numerical scale after augmentation.
    \item Augmentation Application: The final augmentation is applied as a composition \( a_f = t_{f \leftarrow x} \circ a_x \circ t_{f \to x} \), where \( a_x \) is a conventional image-based augmentation (e.g., from AutoAugment or TrivialAugment) operating on the normalized features.
\end{itemize}
This pipeline enables the direct application of standard image augmentations~\cite{AutoAug, TrivialAug} to the frozen feature representations of vision transformers, improving model generalization for downstream tasks.

\subsubsection{L-BFGS optimization for training efficiency}
The L-BFGS algorithm is a popular quasi-Newton method used for solving unconstrained optimization problems~\cite{pb_L-BFGS_ICML2018,sL-BFGS_Trans2019}.
The algorithm exhibits global convergence and enjoys superlinear convergence velocity, attributes highly desirable in optimization contexts requiring accuracy and efficiency.
The key idea behind quasi-Newton methods is to approximate the Hessian matrix of the objective function using only gradient information, which can be computationally more efficient than computing the Hessian directly.
For an optimization problem seeking to minimize $f(x)$, let
\[
    g_k = \nabla f(x_k),~~
    d_k = -B_k g_k,~~
    x_{k+1} = x_k + \alpha_k d_k,
\]
where $\alpha_k$ ensures function value reduction.
The BFGS update for $B_k$, approximating the inverse Hessian, is
\begin{equation}
B_{k+1} = \left(I - \rho_k y_k s_k^\top\right) B_k \left(I - \rho_k s_k y_k^\top\right) + \rho_k s_k s_k^\top
\end{equation}
with $\rho_k = 1 / y_k^\top s_k$, $s_k = x_{k+1} - x_k$, and $y_k = g_{k+1} - g_k$.

This ensures $B_{k+1}$'s positive definiteness and symmetry, which is crucial for the algorithm's stability and efficiency.
The L-BFGS method is known for its numerical stability. It is often preferred over other quasi-Newton methods due to its superior convergence properties.

\subsubsection{LoRA finetuning for parameter efficiency}
Parameter-efficient finetuning methods based on large-scale pretrained models have achieved excellent performance in various downstream applications. One such method, tuning using LoRA~\cite{LoRA_ICLR2022}, introduces an additional low-rank matrix training module to the mapping weights in the multi-head attention layer. Applying this approach to the FiAF task confers significant advantages in parameter efficiency. 

Transitioning from a large parameter space to a low-rank approximation that is capable of introducing new information into the pretrained model is a key aspect of LoRA. 
Kumar et al. hypothesize the updates to the weights during adaptation have a low “intrinsic rank”~\cite{LoRA_ICLR2022}. Given a pretrained weight matrix $ W_0 \in \mathbb{R}^{d \times k}  $ , they constrain the weight update by representing it as a low-rank decomposition: $ W_0 + \Delta W = W_0 + BA $, where $ B \in \mathbb{R}^{d \times r} $ and $ A \in \mathbb{R}^{r \times k} $ , with the rank $ r \ll  min(d, k) $.
During training, $W_0$ is fixed and does not receive gradient updates, while $A$ and $B$ contain trainable parameters. Note both $W_0$ and $ \Delta W = BA $ are multiplied with the same input, and their respective output vectors are summed coordinate-wise. The original transformation is given by $h = W_0x$, and the modified forward pass yields:
\begin{equation}
\label{eq:LoRA}
    h = W_0x+\Delta Wx = W_0x + BAx
\end{equation}
In Eq.~(\ref{eq:LoRA}), $A$ is initialized with random Gaussian values and $B$ is initialized to zero, ensuring that $\Delta W = W_0 + BA $ is zero at the start of training.

\begin{figure}[t]
\centering
\includegraphics[width=1.0\columnwidth]{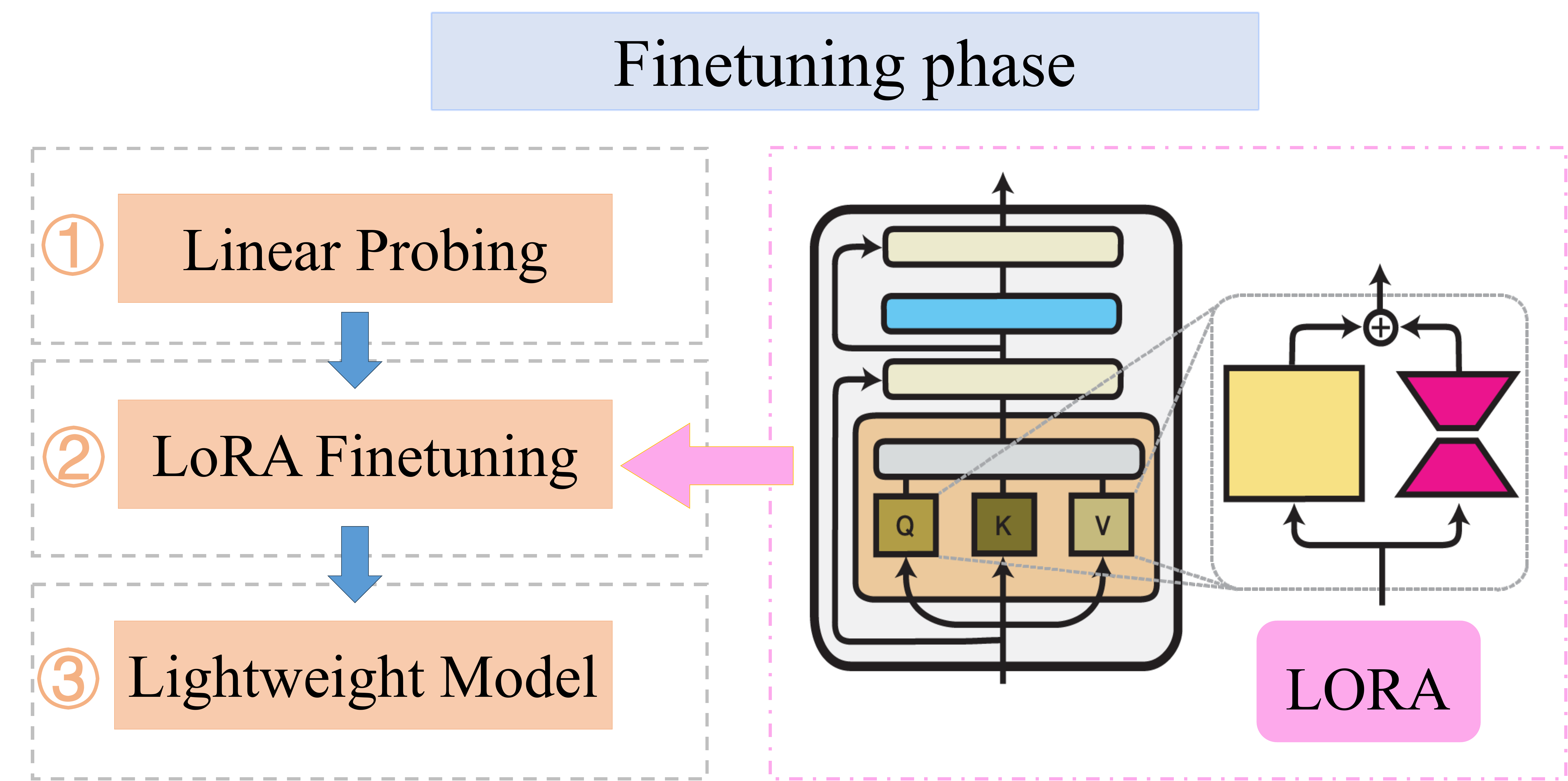}
\caption{$ \mathbf{L^3FACT} $ : \textbf{L}P-\textbf{L}oRA-\textbf{L}M \textbf{F}inetuning in \textbf{ACT}ive finetuning. From the perspective of optimizing parameter efficiency, we propose a variant of the FACT method, $ \mathbf{L^3FACT} $, by replacing the second phase of full finetuning in the finetuning phase shown in Figure~\ref{fig:THriFT} with LoRA.}
\label{fig:ViT-LoRA}
\end{figure}

\subsubsection{FACT and its variant}
%

LP-FT mitigates the issue of distorting pretrained features during full finetuning on large datasets with distribution shifts. We aim to address this with LP-FT in the \textit{FiAF} task, particularly focusing on the challenge of limited samples. In \textit{FiAF}, finetuning is often hindered by overfitting due to over-parameterization, where the small dataset fails to match the pretrained model's parameter complexity. To tackle this, we employ a lightweight model~\cite{MAP_ICML2019} trained on features extracted by the pretrained model.


Our proposed \textbf{FACT} framework is a three-phase, stepwise optimization finetuning approach that involves initial linear probing (LP), followed by full finetuning (FF), and concludes with lightweight model (LM) training on the features of finetuned data. This principled combination, LP-FT for feature preservation plus a lightweight regularizer for overfitting mitigation, distinguishes FACT from heuristic multi-stage pipelines. In addition, we consider replacing LP-FT with LP-LoRA for parameter-efficient finetuning. Replacing LP-FT with LP-LoRA will form the \textbf{L}P-\textbf{L}oRA-\textbf{L}M ($ \mathbf{L^3FACT} $) finetuning approach, which is a favorable variant of our proposed FACT finetuning approach. A more intuitive illustration of the approach's workflow can be found in Figure~\ref{fig:THriFT} and Figure~\ref{fig:ViT-LoRA}. Our finetuning framework addresses the drawbacks of the previously used full finetuning approach, as analyzed below Eq.~\eqref{eq:FF_distort_features}, and improves training and parameter efficiency. Our approach has demonstrated superior performance empirically.


\section{Experiments}
\label{sec:exp}

\subsection{Experimental Setup}
Our experiments were conducted on two RTX 3090 GPUs, using CUDA Toolkit version 11.0 and PyTorch version 1.7.1. Additionally, Scikit-learn version 1.0.2 was employed for the logistic regression experiments.
%

Given the demonstrated rapid efficiency and favorable outcomes, the active finetuning method utilized in our experimental setup is designated as ActiveFT~\cite{ActiveFT_CVPR2023}.

\subsubsection{Datasets and Metrics}
Our experiments utilize seven image classification datasets across three categories, including three classic datasets CIFAR10~\cite{cifar100_2009}, CIFAR100~\cite{cifar100_2009} and ImageNet-1k~\cite{imagenet_IJCV2015}, two imbalanced datasets CIFAR10-LT~\cite{LDAM-DRW_NIPS2019} and CIFAR100-LT~\cite{LDAM-DRW_NIPS2019}, and two fine-grained datasets StanfordCars~\cite{StanfordCars_2013} and FGVCAircraft~\cite{FGVCAircraft_2013}.
The imbalanced datasets were created using the method proposed by LDAM-DRW~\cite{LDAM-DRW_NIPS2019}.
%
%
Top-1 accuracy was used as the evaluation metric in all experiments.

CIFAR10~\cite{cifar100_2009} dataset consists of 60,000 32x32 colour images in 10 classes, with 6,000 images per class. 1,000 images are randomly selected from each category, resulting in a total of 10,000 images forming the test set, and the other 50,000 images are used as the training set.

CIFAR100~\cite{cifar100_2009} dataset is just like the CIFAR10, except it has 100 classes containing 600 images each. There are 500 training images and 100 testing images per class. The 100 classes in the CIFAR-100 are grouped into 20 superclasses.

ImageNet-1k~\cite{imagenet_IJCV2015} dataset includes 1.28 million training images and 50,000 validation images across 1,000 categories. Each category has a variable number of images, making it one of the most comprehensive datasets for image classification tasks.

CIFAR10-LT~\cite{LDAM-DRW_NIPS2019} dataset is an imbalanced version of CIFAR10, created using the method proposed by LDAM-DRW (cite). It consists of 10 classes, with 12,406 training images and 2,478 test images.

CIFAR100-LT~\cite{LDAM-DRW_NIPS2019} dataset is an imbalanced version of CIFAR100, also created using the LDAM-DRW method. It consists of 100 classes, with 8,144 training images and 8,041 test images.

StanfordCars~\cite{StanfordCars_2013} dataset is a fine-grained classification dataset containing 196 classes of cars, with 8,144 training images and 8,041 test images, where each class has been split roughly in a 50-50 split. Classes are typically at the level of Make, Model, Year, ex. 2012 Tesla Model S or 2012 BMW M3 coupe.

FGVCAircraft~\cite{FGVCAircraft_2013} dataset comprises 100 classes of aircraft, with 6,667 training images and 3,333 test images. This dataset is specifically designed for fine-grained visual categorization tasks. This dataset was used as part of the fine-grained recognition challenge FGComp 2013 and it is now a widely used benchmark dataset for fine-grained visual classification of aircraft.

\subsubsection{Architecture Backbone}
We conduct experiments using three representative backbones: ViT~\cite{ViT_ICLR2021}, ViL~\cite{ViL_ICLR25}, and ConvNeXt~\cite{ConvNeXt_CVPR2022} to evaluate the effectiveness of our proposed finetuning framework. They represent the Transformer, xLSTM, and convolutional network architectures, respectively.
Vision Transformer (ViT) demonstrates outstanding performance across various computer vision tasks. Most of our experiments, including those applying the $ \mathrm{L^3FACT} $ method and those on imbalanced and fine-grained datasets, are based on ViT.
Vision LSTM (ViL) adapts the xLSTM building block for computer vision, which has recently gained attention for its exceptional performance.
ConvNeXt is a high-performing backbone that integrates Transformer design elements into the ResNet architecture, achieving competitive results on datasets like ImageNet.
We select ViT-small (22M), ViL2-small (23M), ConvNeXt-T (28M), for our experiments.

\subsubsection{Implementation Details}

The experimental process of this paper can be summarized into two parts: (1) Active sample selection, which selects the most valuable samples through active learning and simulates the labels of the selected samples in the experiment. (2) finetuning, which uses the labeled selected samples to complete the finetuning of the pretrained model through supervised learning.

Feature extraction is a critical step in our experimental setup. We use three different backbone models, each producing 384-dimensional feature vectors for every input image, suitable for both active finetuning and logistic regression classification tasks. For ViT, we modify the architecture by removing the classification head to facilitate feature extraction. A similar approach is applied to ConvNeXt. For ViL, we utilize its built-in 'feature' mode from the official code repository, specifically designed for feature extraction.

This paper focuses on how to perform efficient finetuning based on actively selected samples. In order to ensure the universality of the finetuning strategy proposed in the article, we applied a more general method to complete the sample selection. 
Active sample selection is completed in two parts. First, the image features are extracted through the pretrained encoder, and then the active sample method is applied to find the most helpful samples for model finetuning. In order to simulate a closed finetuning environment, we use the same pretrained model as the finetuning part for feature extraction. In the experiment, we used ViT-small (22M), ViL2-small (23M) and ConvNeXt-T (28M). We deleted the classification head of these models and encoded each image in the dataset into a 384-dimensional feature vector. We refer to the method proposed by ActiveFT \cite{ActiveFT_CVPR2023} to select image subsets, and finally get 0.5\% to 30\% (depending on the dataset) of samples for subsequent finetuning.

The finetuning process consists of three basic modules FT, LP, LoRA and their combinations (e.g. LP-FT). 
%
All images in the experiments are processed at a resolution of \(224 \times 224\) pixels. For full finetuning, we trained for $300$ epochs, with evaluations every $6$ epochs. Other finetuning modes were trained for $50$ epochs, with evaluations after each epoch to maintain consistency. 
We use a cosine learning rate schedule.
While the cosine learning rate scheduler is commonly initialized with a base learning rate of 0.01, we found that this configuration led to training instability when applied to ConvNeXt architectures. Empirical observations suggested that increasing the initial learning rate to 0.1 improved training convergence and stability.
This section details the Frozen Feature Augmentation (FroFA~\cite{FroFA_CVPR2024}) technique and its variants. The $\mathrm{Default~FroFA}$ applies global augmentation with shared random values across channels (e.g., for contrast scaling), while $\mathrm{Channel~FroFA~(cFroFA)}$ introduces channel-independent parameters for increased stochasticity. The most effective variant, $\mathrm{Channel^2~FroFA~(c_2FroFA)}$, extends $\mathrm{cFroFA}$ by incorporating channel-wise min/max statistics, particularly improving brightness augmentation stability. Following B\"ar et al.~\cite{FroFA_CVPR2024}'s findings on $\mathrm{c_2FroFA}$'s superior performance, our implementation specifically uses this variant for brightness augmentation, referred to throughout as $\mathrm{FroFA}$.

\definecolor{mylightblue}{RGB}{204,229,255}
\definecolor{mylightgreen}{RGB}{224, 246, 216}
\begin{table*}[t]
\centering
\caption{Classifcation accuracy comparisons on CIFAR10, CIFAR100 and ImageNet-1k with different sampling ratios.}
\resizebox{1.0\textwidth}{!}{
\begin{threeparttable}
\begin{tabular}{l|ccccc|cccc|ccc}
\hline
\multirow{2}{*}{Methods} & \multicolumn{5}{c|}{CIFAR10}          & \multicolumn{4}{c|}{CIFAR100} & \multicolumn{3}{c}{ImageNet-1k} \\
                         & 0.1\% & 0.2\% & 0.5\% & 1\%   & 2\%   & 1\%   & 2\%   & 5\%   & 10\%  & 0.5\%     & 1\%      & 2\%      \\ \hline
ActiveFT~\cite{ActiveFT_CVPR2023} & - & - & 85.00 & 88.20 & 90.10 & 26.10 & 40.70 & 54.60 & 71.00 & 36.80 & 50.10 & 54.20 \\
BiLAF~\cite{BiLAF_NIPS24} & - & - & 81.00 & 89.20 & 92.50 & 31.80 & 43.50 & 62.80 & 73.70 & - & 50.80 & 56.90 \\
ActiveDC~\cite{xu2024activedc} & 61.30 & 73.10 & 87.30 & 88.90 & 90.30 & 34.50 & 54.60 & 71.90 & 74.30 & 50.90 & 56.30 & 60.10 \\
VeCAF~\cite{VeCAF_MM24}\textsubscript{multi-run} & - & - & - & \textbf{93.57} & \textbf{95.27} & - & - & - & - & - & 58.31 & 63.76 \\
\rowcolor{mylightgreen}
FF                       & 54.39 & 71.86 & 85.00 & 88.20 & 90.10 & 26.10 & 40.70 & 54.60 & 71.00 & 36.80     & 50.10    & 54.20    \\
LP                       & 76.30 & 82.71 & 86.35 & 86.81 & 90.46 & 23.95 & 51.16 & 65.36 & 71.57 & 53.01     & 59.03    & 63.74    \\
LP-FT                    & 76.30 & 83.20 & 88.34 & 89.27 & 93.38 & 37.47 & 59.16 & 71.92 & 78.35 & 53.67     & 59.63    & 64.18    \\
LP-LoRA                  & 76.41 & 84.60 & 88.18 & 89.96 & 93.21 & 42.04 & 59.17 & 69.48 & 75.52 & 53.06     & 59.20    & 63.74    \\ \hline
\rowcolor{mylightblue}
FACT~(ours)               & 87.13 & 89.76 & 90.16 & 92.22 & 94.05 & 61.09 & 70.54 & 76.79 & 80.66 & 57.58     & 61.95    & 65.33    \\
                         & \textbf{\textcolor{Red}{$\uparrow\Delta$32.74}} \tnote{*} & \textbf{\textcolor{Red}{$\uparrow\Delta$17.90}} & \textbf{\textcolor{Red}{$\uparrow\Delta$5.16}}  & \textbf{\textcolor{Red}{$\uparrow\Delta$4.02}}  & \textbf{\textcolor{Red}{$\uparrow\Delta$3.95}}  & \textbf{\textcolor{Red}{$\uparrow\Delta$34.99}} & \textbf{\textcolor{Red}{$\uparrow\Delta$29.84}} & \textbf{\textcolor{Red}{$\uparrow\Delta$22.19}} & \textbf{\textcolor{Red}{$\uparrow\Delta$9.66}}  & \textbf{\textcolor{Red}{$\uparrow\Delta$20.78}} & \textbf{\textcolor{Red}{$\uparrow\Delta$11.85}} & \textbf{\textcolor{Red}{$\uparrow\Delta$11.13}}  \\
\rowcolor{mylightblue}
$ \mathrm{L^3FACT} $~(ours)  & 87.04 & 90.35 & 90.79 & 92.76 & 94.25 & 64.58 & 72.12 & 76.17 & 78.78 & 58.69     & 62.41    & 65.73    \\
                         & \textbf{\textcolor{Red}{$\uparrow\Delta$32.65}} & \textbf{\textcolor{Red}{$\uparrow\Delta$18.49}} & \textbf{\textcolor{Red}{$\uparrow\Delta$5.79}}  & \textbf{\textcolor{Red}{$\uparrow\Delta$4.56}}  & \textbf{\textcolor{Red}{$\uparrow\Delta$4.15}}  & \textbf{\textcolor{Red}{$\uparrow\Delta$38.48}} & \textbf{\textcolor{Red}{$\uparrow\Delta$31.42}} & \textbf{\textcolor{Red}{$\uparrow\Delta$21.57}} & \textbf{\textcolor{Red}{$\uparrow\Delta$7.78}}  & \textbf{\textcolor{Red}{$\uparrow\Delta$21.89}} & \textbf{\textcolor{Red}{$\uparrow\Delta$12.31}}  & \textbf{\textcolor{Red}{$\uparrow\Delta$11.53}}  \\ \hline
\end{tabular}
\begin{tablenotes}
\item[*] The red font indicates the improvement in accuracy achieved by our finetuning framework compared to the full finetuning. FACT's strength is a further refinement of LP-FT.
\end{tablenotes}
\end{threeparttable}
}
\label{tab:data_1}
\end{table*}

\begin{table*}[t]
\centering
\caption{Classifcation accuracy comparisons on long-tail datasets (CIFAR10-LT and CIFAR100-LT).}
\resizebox{0.65\textwidth}{!}{
\begin{tabular}{l|ccc|cccc}
\hline
\multirow{2}{*}{Methods} & \multicolumn{3}{c|}{CIFAR10-LT} & \multicolumn{4}{c}{CIFAR100-LT} \\
                         & 0.5\%     & 1\%      & 2\%      & 5\%    & 10\%   & 20\%  & 30\%  \\ \hline
FF                       & 67.31     & 79.02    & 87.21    & 46.69  & 58.61  & 74.05 & 79.96 \\
LP                       & 84.38     & 87.25    & 89.02    & 59.92  & 66.40  & 73.11 & 77.05 \\
LP-FT                    & 84.38     & 87.29    & 89.43    & 63.44  & 71.05  & 78.60 & 81.79 \\
LP-LoRA                  & 84.38     & 87.29    & 89.35    & 63.30  & 70.48  & 77.99 & 79.59 \\ \hline
\multirow{2}{*}{FACT}    & 90.92     & 91.93    & 91.69    & 70.58  & 77.15  & 82.68 & 84.05 \\
                         & \textbf{\textcolor{Red}{$\uparrow\Delta$23.61}}     & \textbf{\textcolor{Red}{$\uparrow\Delta$12.91}}    & \textbf{\textcolor{Red}{$\uparrow\Delta$4.48}}     & \textbf{\textcolor{Red}{$\uparrow\Delta$23.89}}  & \textbf{\textcolor{Red}{$\uparrow\Delta$18.54}}  & \textbf{\textcolor{Red}{$\uparrow\Delta$8.63}}  & \textbf{\textcolor{Red}{$\uparrow\Delta$4.09}}  \\
\multirow{2}{*}{$ \mathrm{L^3FACT} $}     & 90.92     & 91.93    & 93.18    & 71.75  & 77.24  & 81.09 & 82.50 \\
                         & \textbf{\textcolor{Red}{$\uparrow\Delta$23.61}}     & \textbf{\textcolor{Red}{$\uparrow\Delta$12.91}}    & \textbf{\textcolor{Red}{$\uparrow\Delta$5.97}}     & \textbf{\textcolor{Red}{$\uparrow\Delta$25.06}}  & \textbf{\textcolor{Red}{$\uparrow\Delta$18.63}}  & \textbf{\textcolor{Red}{$\uparrow\Delta$7.04}}  & \textbf{\textcolor{Red}{$\uparrow\Delta$2.54}}  \\ \hline
\end{tabular}
}
\label{tab:data_2}
\end{table*}

\begin{table*}[t]
\centering
\caption{Classifcation accuracy comparisons on fine-grained datasets (StanfordCars and FGVCAircraft).}
\resizebox{0.75\textwidth}{!}{
\begin{tabular}{l|cccc|cccc}
\hline
\multirow{2}{*}{Methods} & \multicolumn{4}{c|}{Stanford-Cars} & \multicolumn{4}{c}{FGVCAircraft} \\
                         & 5\%     & 10\%   & 20\%   & 30\%   & 5\%    & 10\%   & 20\%   & 30\%  \\ \hline
FF                       & 6.82    & 15.18  & 35.39  & 51.87  & 14.43  & 23.55  & 33.93  & 47.34 \\
LP                       & 4.34    & 9.00   & 16.52  & 23.55  & 9.36   & 16.95  & 27.30  & 33.24 \\
LP-FT                    & 7.45    & 15.33  & 32.94  & 48.92  & 14.67  & 24.33  & 37.59  & 47.88 \\
LP-LoRA                  & 7.97    & 16.45  & 32.11  & 44.98  & 15.63  & 26.64  & 37.11  & 46.29 \\ \hline
\multirow{2}{*}{FACT}    & 13.17   & 23.02  & 42.16  & 57.31  & 19.83  & 29.16  & 41.64  & 51.79 \\
                         & \textbf{\textcolor{Red}{$\uparrow\Delta$6.35}}    & \textbf{\textcolor{Red}{$\uparrow\Delta$7.84}}   & \textbf{\textcolor{Red}{$\uparrow\Delta$6.77}}   & \textbf{\textcolor{Red}{$\uparrow\Delta$5.44}}   & \textbf{\textcolor{Red}{$\uparrow\Delta$5.40}}   & \textbf{\textcolor{Red}{$\uparrow\Delta$5.61}}   & \textbf{\textcolor{Red}{$\uparrow\Delta$7.71}}   & \textbf{\textcolor{Red}{$\uparrow\Delta$4.45}}  \\
\multirow{2}{*}{$ \mathrm{L^3FACT} $}     & 14.50   & 24.72  & 41.65  & 53.38  & 23.91  & 33.99  & 43.29  & 51.79 \\
                         & \textbf{\textcolor{Red}{$\uparrow\Delta$7.68}}    & \textbf{\textcolor{Red}{$\uparrow\Delta$9.54}}   & \textbf{\textcolor{Red}{$\uparrow\Delta$6.26}}   & \textbf{\textcolor{Red}{$\uparrow\Delta$1.51}}   & \textbf{\textcolor{Red}{$\uparrow\Delta$9.48}}   & \textbf{\textcolor{Red}{$\uparrow\Delta$10.44}}  & \textbf{\textcolor{Red}{$\uparrow\Delta$9.36}}   & \textbf{\textcolor{Red}{$\uparrow\Delta$4.45}}  \\ \hline
\end{tabular}
}
\label{tab:data_3}
\end{table*}

%



\begin{figure*}[t]
\centering
\includegraphics[width=1.0\textwidth]{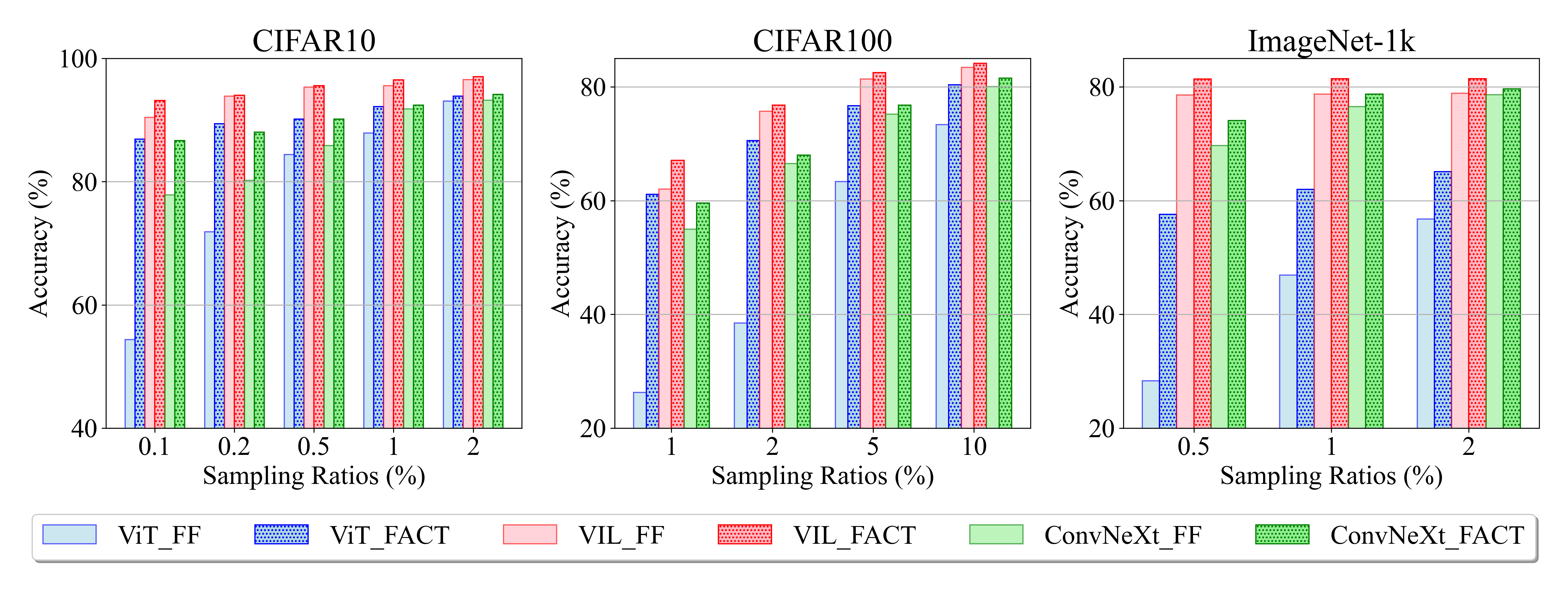}
\caption{Classification accuracy of methods applied on different backbones on CIFAR10, CIFAR100, and ImageNet-1k.}
\label{fig:tab2fig_backbones}
\end{figure*}

\subsection{Benchmark Results}
%
%
Our finetuning frameworks for the \textit{FiAF} task, FACT and $\mathrm{L^3FACT}$, were experimentally compared with basic baselines, including full finetuning of ActiveFT~\cite{ActiveFT_CVPR2023}, ActiveDC~\cite{xu2024activedc}, BiLAF~\cite{BiLAF_NIPS24}, VeCAF~\cite{VeCAF_MM24} and LP, LP-FT, and LP-LoRA.
Our experimental section is structured into three main components:

\subsubsection{Comparison on Classic Image Datasets}
We have conducted an evaluation of various finetuning methods applied to the ViT backbone across three classic image classification datasets, using different sampling ratios. It ensures comprehensive and comparative insights into the performance of these methods. The results of these experiments are summarized in Table~\ref{tab:data_1}.
Our experiments reveal that the FF method exhibits suboptimal performance, particularly when the sampling ratio is low. In contrast, the LP approach demonstrates superior results on small-scale finetuning datasets. Interestingly, as we increase the amount of finetuning data, the performance of FF improves significantly, eventually reaching parity with LP. The LP-FT and LP-LoRA methods exhibit similar performance, both being superior to LP and FF. Our proposed frameworks, FACT and $ \mathrm{L^3FACT} $, achieve better results across different sampling ratios of multiple datasets.

In addition, we evaluated our method against existing state-of-the-art approaches in active finetuning. Experimental results indicate that our framework consistently outperforms prior works, achieving substantial gains in performance. This advancement represents a notable step forward in active finetuning research. 
Notably, the VeCAF method demonstrates competitive performance, achieving the highest accuracy on the CIFAR-10 dataset (bold in Table~\ref{tab:data_1}). This performance gain can be attributed to its innovative integration of language-embedded knowledge for feature enhancement and its multi-round data selection strategy.

In our results presentation in Table~\ref{tab:data_1}, \colorbox{mylightgreen}{light-green} shaded entries indicate performance metrics obtained through full finetuning using the ActiveFT data selection method, while \colorbox{mylightblue}{light-blue} shaded entries represent the outcomes of our proposed finetuning framework. The \textcolor{red}{red-highlighted} values quantify the accuracy improvements achieved by our approach relative to the baseline.

\subsubsection{Evaluation on Sample-Imbalanced and Fine-Grained Datasets}

To evaluate the robustness and generalization of our proposed finetuning method, we introduce two sample-imbalanced datasets and two fine-grained classification benchmarks (detailed in ``Datasets and Metrics''). Experimental results (Tables~\ref{tab:data_2} and~\ref{tab:data_3}) demonstrate that our method achieves consistent performance on imbalanced datasets, aligning with evaluations on classical image classification tasks. In contrast, LP exhibits significant limitations on fine-grained datasets compared to FF. This discrepancy arises because pretrained generic features exhibit limited discriminative power for fine-grained visual distinctions, necessitating adaptive optimization of the backbone parameters. Remarkably, our method outperforms existing approaches in both scenarios.
Furthermore, the framework maintains stable performance gains across varying sampling ratios, consistent with observations from classic image classification benchmarks. Critically, it demonstrates superior generalization capabilities under both imbalanced data distributions and fine-grained recognition tasks, highlighting its robustness to diverse challenges.

\subsubsection{Analysis across Diverse Network Architectures}
We further investigate the effectiveness of our finetuning framework by applying it to three different pretrained models: ViT, ViL, and ConvNeXt, all pretrained on ImageNet-1K. 
%
%
These experiments are conducted on the same three classic image classification datasets mentioned earlier. The findings from this analysis are illustrated in Figure~\ref{fig:tab2fig_backbones}.
It presents comparative histograms differentiated by distinct shading patterns and colors, where ViT\_FF denotes full finetuning of ViT architectures, while ViL\_FACT indicates our proposed FACT method implemented on ViL architectures. The remaining figure legends follow analogous naming conventions to distinguish between baseline architectures and their corresponding finetuning strategies.

We compare the FF method and our FACT method under different backbones in the \textit{FiAF} task. The results indicate that, across architectures including ViT, ConvNext, and ViL, FACT consistently outperforms FF, particularly when finetuning with limited data samples. Therefore, our finetuning framework is universally applicable to different architectural models. As the sampling ratio increases, the performance of both FACT and FF converges, with both methods achieving strong results.
Hence, FACT is most beneficial under extreme low-budget regimes or severe distribution shift or class imbalance.

%
Additionally, we observe that in the \textit{FiAF} task, ViL demonstrates strong performance across different sampling ratios, significantly outperforming ViT. 
%
We believe ViL outperforms ViT in our tasks due to its stronger inductive biases: the stateful memory in mLSTM introduces a regularization effect through gating mechanisms, which helps prevent overfitting and captures spatial dependencies more effectively. Meanwhile, the implicit positional encoding enhances resolution generalization by eliminating reliance on fixed grid-based embeddings, thus maintaining robustness when adapting to new resolutions or data distributions. Additionally, local convolutional layers help preserve fine-grained local details, while bidirectional sequence modeling provides implicit data augmentation by learning from complementary spatial perspectives. Together, these mechanisms make the model more robust in few-shot finetuning scenarios compared to purely data-driven attention.

\begin{table}[t]
\centering
\caption{Comparison of training efficiency (minute).}
\resizebox{1.0\columnwidth}{!}{
\begin{tabular}{lcccccc}
\hline
\multirow{2}{*}{datasets} & \multicolumn{2}{c}{ViT} & \multicolumn{2}{c}{ViL} & \multicolumn{2}{c}{ConvNeXt} \\
                          & FF        & FACT      & FF        & FACT      & FF          & FACT         \\ \hline
CIFAR10 (2\%)              & 13        & 5           & 120       & 26          & 14          & 5              \\
CIFAR100 (10\%)            & 51        & 13          & 316       & 65          & 56          & 15             \\
ImageNet-1k (2\%)             & 245       & 66          & 1636      & 512         & 271         & 208            \\ \hline
\end{tabular}
}
\label{tab:efficiency}
\end{table}

\subsection{Efficiency Analysis}

%
%
Our method not only achieves significant breakthroughs in performance but also demonstrates efficiency in training time and parameter usage. As illustrated in Table~\ref{tab:efficiency}, we compared the training durations of the FF method and our FACT method under different backbones. Due to the pre-operation of linear probing and the application of a quasi-Newton optimizer, the convergence of our model is notably accelerated. The training time has decreased on multiple datasets, thereby enhancing training efficiency.
Despite demonstrating superior accuracy (Figure~\ref{fig:tab2fig_backbones}), ViL incurs significantly higher computational latency compared to ViT and ConvNeXt at equivalent parameter scales, as quantified in Table~\ref{tab:efficiency}.
In addition, the variant of the FACT framework, $ \mathrm{L^3FACT} $ , significantly reduces the number of trainable parameters from 21,669,514 to 151,306 compared to FF (a reduction to \textbf{only 6.98‰} of the original count). Thus, the finetuning framework we applied to the \textit{FiAF} task is a parameter-efficient finetuning framework.


\subsection{Further General Capability Analysis}

\begin{table*}[t]
\centering
\caption{Comparison of different model scales and pretraining methods}
\begin{tabular}{lll|ccccc|cccc}
\hline
\multirow{2}{*}{pretrain} & \multirow{2}{*}{models} & \multirow{2}{*}{methods} & \multicolumn{5}{c|}{CIFAR10} & \multicolumn{4}{c}{CIFAR100} \\
                          &                         &                          & 0.1\% & 0.2\% & 0.5\% & 1\% & 2\% & 1\% & 2\% & 5\% & 10\% \\ \hline
\multirow{4}{*}{DINO~\cite{DINO_ICCV2021}} & \multirow{2}{*}{vit-small} & FF   & \cellcolor{gray!10}54.39 & \cellcolor{gray!10}71.86 & \cellcolor{gray!10}85.00 & \cellcolor{gray!10}88.20 & \cellcolor{gray!10}90.10 & \cellcolor{gray!10}26.10 & \cellcolor{gray!10}40.70 & \cellcolor{gray!10}54.60 & \cellcolor{gray!10}71.00 \\
& & FACT & \cellcolor{gray!10}87.13 & \cellcolor{gray!10}89.76 & \cellcolor{gray!10}90.16 & \cellcolor{gray!10}92.22 & \cellcolor{gray!10}94.05 & \cellcolor{gray!10}61.09 & \cellcolor{gray!10}70.54 & \cellcolor{gray!10}76.79 & \cellcolor{gray!10}80.66 \\
& \multirow{2}{*}{vit-base} & FF   & 68.42 & 83.53 & 90.98 & 93.90 & 96.28 & 36.11 & 60.45 & 75.61 & 82.05 \\
& & FACT & 82.91 & 92.12 & 94.37 & 96.10 & 96.69 & 54.46 & 72.54 & 79.82 & 83.75 \\ \hline
\multirow{4}{*}{CLIP~\cite{CLIP_ICML2021}} & \multirow{2}{*}{vit-base} & FF   & \cellcolor{gray!10}72.32 & \cellcolor{gray!10}83.06 & \cellcolor{gray!10}91.74 & \cellcolor{gray!10}93.02 & \cellcolor{gray!10}93.82 & \cellcolor{gray!10}59.36 & \cellcolor{gray!10}67.05 & \cellcolor{gray!10}74.48 & \cellcolor{gray!10}78.11 \\
& & FACT & \cellcolor{gray!10}75.97 & \cellcolor{gray!10}88.86 & \cellcolor{gray!10}92.09 & \cellcolor{gray!10}93.57 & \cellcolor{gray!10}94.25 & \cellcolor{gray!10}61.43 & \cellcolor{gray!10}69.27 & \cellcolor{gray!10}75.27 & \cellcolor{gray!10}78.39 \\
& \multirow{2}{*}{vit-large} & FF   & 50.12 & 71.57 & 83.47 & 89.19 & 92.40 & 37.67 & 53.89 & 68.57 & 76.68 \\
& & FACT & 55.30 & 75.29 & 85.56 & 91.06 & 93.24 & 45.73 & 61.31 & 72.76 & 77.74 \\ \hline
\end{tabular}
\label{tab:rq_pretrain}
\end{table*}

\begin{table*}[t]
\centering
\caption{Comparison of Parameter-Efficient Fine-Tuning Methods}
\begin{tabular}{l|ccccc|cccc|c}
\hline
\multirow{2}{*}{Methods} & \multicolumn{5}{c|}{CIFAR10}          & \multicolumn{4}{c|}{CIFAR100}  & \multirow{2}{*}{Trainable} \\
                         & 0.1\% & 0.2\% & 0.5\% & 1.0\% & 2.0\% & 1.0\% & 2.0\% & 5.0\% & 10.0\% &                            \\ \hline
FF                       & 54.39 & 71.86 & 85.00 & 88.20 & 90.10 & 26.10 & 40.70 & 54.60 & 71.00  & 100\%                      \\
FACT                     & \textbf{87.13} & \ul{89.76} & 90.16 & \ul{92.22} & \ul{94.05} & 61.09 & \ul{70.54} & \textbf{76.79} & \textbf{80.66}  & 100\%                      \\ \hline
$\mathrm{FACT_{lora}}$   & \ul{87.04} & \textbf{90.35} & \ul{90.79} & \textbf{92.76} & \textbf{94.25} & \textbf{64.58} & \textbf{72.12} & \ul{76.17} & \ul{78.78}  & 0.69\%                     \\
$\mathrm{FACT_{prompt}}$ & 83.34 & 87.12 & \textbf{90.97} & 92.11 & 92.71 & \ul{62.59} & 68.85 & 73.69 & 76.44  & 6.19\%                     \\
$\mathrm{FACT_{prefix}}$ & 82.72 & 86.47 & 89.10 & 90.47 & 90.38 & 58.62 & 64.19 & 69.16 & 72.00  & 0.65\%                     \\ \hline
\end{tabular}
\label{tab:rq_peft}
\end{table*}

\begin{table*}[t]
\centering
\caption{Comparison of semantic segmentation mIoU and accuracy on ADE20k dataset}
\begin{tabular}{l|ccc|ccc}
\hline
     & \multicolumn{3}{c|}{ADE20k (5\%)}                & \multicolumn{3}{c}{ADE20k (10\%)}                \\
     & mIoU (\%)      & mAcc (\%)      & aAcc (\%)      & mIoU (\%)      & mAcc (\%)      & aAcc (\%)      \\ \hline
FF   & 17.61          & 24.27          & 66.58          & 23.05          & 30.77          & 71.40          \\
LP   & 18.53          & 23.06          & 69.46          & 23.53          & 30.80          & 71.60          \\
FACT & \textbf{22.39} & \textbf{29.45} & \textbf{71.16} & \textbf{24.87} & \textbf{32.20} & \textbf{72.89} \\ \hline
\end{tabular}
\label{tab:rq_mmseg}
\end{table*}

\subsubsection{Comparison of Model Scales and Pretraining Strategies}

To further validate the generalizability of our proposed method, we conduct extensive experiments across diverse pre-training strategies (specifically DINO~\cite{DINO_ICCV2021} and CLIP~\cite{CLIP_ICML2021}) and varying model scales, ranging from ViT-small to ViT-large. As presented in Table~\ref{tab:rq_pretrain}, while performance naturally varies depending on the pretraining source and model capacity, our FACT method consistently outperforms the Full Finetuning (FF) baseline across the vast majority of settings. Regardless of whether the model is initialized with self-supervised DINO or vision-language CLIP weights, FACT demonstrates superior adaptability. This consistent advantage highlights the robustness of our approach, confirming that its effectiveness is not confined to a specific architecture or initialization but holds true even when scaling to larger models and switching between distinct pretraining paradigms.

\subsubsection{Comparison with More Modern PEFT Techniques}

To further validate the versatility and efficiency of our proposed method, we conducted a comprehensive comparison incorporating several representative and widely-adopted Parameter-Efficient Fine-Tuning (PEFT) techniques, including LoRA~\cite{LoRA_ICLR2022}, Visual Prompt Tuning~\cite{VPT_ECCV2022}, and Prefix Tuning~\cite{PrefixTuning_ACL2021}. As detailed in Table~\ref{tab:rq_peft}, the choice of PEFT strategy significantly influences downstream performance. Notably, $\mathrm{FACT_{lora}}$ demonstrates a distinct advantage in terms of the trade-off between parameter efficiency and accuracy. With only 0.69\% of trainable parameters, $\mathrm{FACT_{lora}}$ consistently outperforms $\mathrm{FACT_{prompt}}$ (which requires 6.19\% parameters) and $\mathrm{FACT_{prefix}}$ across various data regimes. For instance, in the CIFAR100 1\% setting, $\mathrm{FACT_{lora}}$ achieves 64.58\% accuracy, surpassing the 62.59\% of prompt tuning and the 58.62\% of prefix tuning. These results indicate that while different PEFT methods impact performance to varying degrees, LoRA offers superior parameter efficiency and effectiveness within the FACT framework.

\subsubsection{Comparison on Semantic Segmentation Tasks}

Having established the effectiveness of our approach on classic, fine-grained, and long-tailed image classification tasks, we extend our evaluation to the dense prediction task of semantic segmentation to further demonstrate its generalizability. The experimental framework adopts MMSegmentation~\cite{mmseg2020}, and the segmentation task head employs UPerNet~\cite{UperNet_ECCV18}. Specifically, we conduct experiments on the ADE20k~\cite{ADE20k_CVPR17} dataset utilizing a ViT-small backbone pretrained with DINO. As presented in Table~\ref{tab:rq_mmseg}, our FACT method consistently surpasses the Full Finetuning (FF) baseline across all evaluation metrics, including mIoU, mAcc, and aAcc. For instance, with only 5\% of the training data, FACT achieves a remarkable mIoU of 22.39\%, outperforming FF by a significant margin of 4.78\%. This performance advantage is maintained at the 10\% data setting, where FACT secures 24.87\% mIoU compared to 23.05\% for FF. These results provide compelling evidence for the robustness and broad applicability of our method across diverse computer vision tasks.


\subsection{Ablation Studies}

\begin{table*}[t]
\centering
\caption{The effect of the FroFA module on FACT and its variants in the finetuning framework.}
\resizebox{0.95\textwidth}{!}{
\begin{tabular}{ll|ccccc|cccc|ccc}
\hline
\multirow{2}{*}{Methods} & \multirow{2}{*}{FroFA} & \multicolumn{5}{c|}{CIFAR10}          & \multicolumn{4}{c|}{CIFAR100} & \multicolumn{3}{c}{ImageNet-1k} \\
                         &                        & 0.1\% & 0.2\% & 0.5\% & 1\%   & 2\%   & 1\%   & 2\%   & 5\%   & 10\%  & 0.5\%     & 1\%      & 2\%      \\ \hline
\multirow{2}{*}{FACT}    & w.o.                   & 86.44 & 89.49 & 90.00 & 92.07 & 93.72 & 60.76 & 70.72 & 76.54 & 80.22 & 57.59     & 61.93    & 64.88    \\
                         & w.                     & \cellcolor{mylightblue}{87.13} & \cellcolor{mylightblue}{89.76} & \cellcolor{mylightblue}{90.16} & \cellcolor{mylightblue}{92.22} & \cellcolor{mylightblue}{94.05} & \cellcolor{mylightblue}{61.09} & \cellcolor{mylightgreen}{70.54} & \cellcolor{mylightblue}{76.79} & \cellcolor{mylightblue}{80.66} & \cellcolor{mylightgreen}{57.58}     & \cellcolor{mylightblue}{61.95}    & \cellcolor{mylightblue}{65.33}    \\
\multirow{2}{*}{$ \mathrm{L^3FACT} $}  & w.o.                   & 86.35 & 90.27 & 90.62 & 92.74 & 94.12 & 64.24 & 71.65 & 75.27 & 77.76 & 57.89     & 62.17    & 63.95    \\
                         & w.                     & \cellcolor{mylightblue}{87.04} & \cellcolor{mylightblue}{90.35} & \cellcolor{mylightblue}{90.79} & \cellcolor{mylightblue}{92.76} & \cellcolor{mylightblue}{94.25} & \cellcolor{mylightblue}{64.58} & \cellcolor{mylightblue}{72.12} & \cellcolor{mylightblue}{76.17} & \cellcolor{mylightblue}{78.78} & \cellcolor{mylightblue}{58.69}     & \cellcolor{mylightblue}{62.41}    & \cellcolor{mylightblue}{65.73}    \\ \hline
\end{tabular}
}
\label{tab:influ_frofa}
\end{table*}

\begin{table*}[t]
\centering
\caption{The effect of the each increment step on FACT and its variants.}
\resizebox{0.95\textwidth}{!}{
\begin{tabular}{l|rrrrr|rrrr|rrr}
\hline
\multirow{2}{*}{Inc of methods} & \multicolumn{5}{c|}{CIFAR10} & \multicolumn{4}{c|}{CIFAR100} & \multicolumn{3}{c}{ImageNet-1k} \\
& 0.1\% & 0.2\% & 0.5\% & 1\% & 2\% & 1\% & 2\% & 5\% & 10\% & 0.5\% & 1\% & 2\% \\ \hline
FF & \cellcolor{blue!5}{0} & \cellcolor{blue!5}{0} & \cellcolor{blue!5}{0} & \cellcolor{blue!5}{0} & \cellcolor{blue!5}{0} 
& \cellcolor{blue!5}{0} & \cellcolor{blue!5}{0} & \cellcolor{blue!5}{0} & \cellcolor{blue!5}{0} 
& \cellcolor{blue!5}{0} & \cellcolor{blue!5}{0} & \cellcolor{blue!5}{0} \\
LP & \cellcolor{blue!50}{21.91} & \cellcolor{blue!42}{10.85} & \cellcolor{blue!17}{1.35} & \cellcolor{green!14}{-1.36} & \cellcolor{blue!7}{0.36} 
& \cellcolor{green!5}{-2.15} & \cellcolor{blue!25}{10.46} & \cellcolor{blue!37}{10.76} & \cellcolor{blue!5}{0.57} 
& \cellcolor{blue!95}{16.21} & \cellcolor{blue!63}{8.93} & \cellcolor{blue!73}{9.54} \\ \hline
LP-FT & \cellcolor{blue!5}{0.00} & \cellcolor{blue!5}{0.49} & \cellcolor{blue!25}{1.99} & \cellcolor{blue!25}{2.46} & \cellcolor{blue!60}{2.92} 
& \cellcolor{blue!34}{13.52} & \cellcolor{blue!19}{8.00} & \cellcolor{blue!23}{6.56} & \cellcolor{blue!45}{6.78} 
& \cellcolor{blue!7}{0.66} & \cellcolor{blue!5}{0.60} & \cellcolor{blue!5}{0.44} \\
FACT\textsubscript{w.o.FroFA} & \cellcolor{blue!23}{10.14} & \cellcolor{blue!25}{6.29} & \cellcolor{blue!21}{1.66} & \cellcolor{blue!28}{2.80} & \cellcolor{blue!7}{0.34} 
& \cellcolor{blue!58}{23.29} & \cellcolor{blue!28}{11.56} & \cellcolor{blue!16}{4.62} & \cellcolor{blue!12}{1.87} 
& \cellcolor{blue!41}{3.92} & \cellcolor{blue!16}{2.30} & \cellcolor{blue!5}{0.70} \\
FACT & \cellcolor{blue!5}{0.69} & \cellcolor{blue!5}{0.27} & \cellcolor{blue!5}{0.16} & \cellcolor{blue!5}{0.15} & \cellcolor{blue!7}{0.33} 
& \cellcolor{blue!5}{0.33} & \cellcolor{green!5}{-0.18} & \cellcolor{blue!5}{0.25} & \cellcolor{blue!5}{0.44} 
& \cellcolor{green!5}{-0.01} & \cellcolor{blue!5}{0.02} & \cellcolor{blue!5}{0.45} \\ \hline
LP-LoRA & \cellcolor{blue!5}{0.11} & \cellcolor{blue!7}{1.89} & \cellcolor{blue!23}{1.83} & \cellcolor{blue!31}{3.15} & \cellcolor{blue!57}{2.75} 
& \cellcolor{blue!45}{18.09} & \cellcolor{blue!19}{8.01} & \cellcolor{blue!14}{4.12} & \cellcolor{blue!26}{3.95} 
& \cellcolor{blue!5}{0.05} & \cellcolor{blue!5}{0.17} & \cellcolor{blue!5}{0.00} \\
$\mathrm{L^3FACT}$\textsubscript{w.o.FroFA} & \cellcolor{blue!23}{9.94} & \cellcolor{blue!22}{5.67} & \cellcolor{blue!31}{2.44} & \cellcolor{blue!28}{2.78} & \cellcolor{blue!19}{0.91} 
& \cellcolor{blue!55}{22.20} & \cellcolor{blue!30}{12.48} &\cellcolor{blue!20}{5.79} & \cellcolor{blue!15}{2.24} 
& \cellcolor{blue!51}{4.83} & \cellcolor{blue!21}{2.97} & \cellcolor{blue!5}{0.21} \\
$\mathrm{L^3FACT}$ & \cellcolor{blue!5}{0.69} & \cellcolor{blue!5}{0.08} & \cellcolor{blue!5}{0.17} & \cellcolor{blue!5}{0.02} & \cellcolor{blue!5}{0.13} 
& \cellcolor{blue!5}{0.34} & \cellcolor{blue!5}{0.47} & \cellcolor{blue!5}{0.90} & \cellcolor{blue!7}{1.02} 
& \cellcolor{blue!8}{0.80} & \cellcolor{blue!5}{0.24} & \cellcolor{blue!14}{1.78} \\ \hline
\end{tabular}
}
\label{tab:contribution_each}
\end{table*}

\subsubsection{The effect of FroFA module}
%
%
%
%
To evaluate the contribution of the frozen feature augmentation (FroFA) module within our finetuning framework (FACT and its variants), we perform ablation studies to analyze its effect on overall performance. In Table~\ref{tab:influ_frofa}, ``w.o.'' indicates the finetuning framework without the FroFA module, while ``w.'' denotes the framework with the FroFA module integrated. Experimental results are highlighted with \colorbox{mylightblue}{light-blue} backgrounds to signify performance improvements and \colorbox{mylightgreen}{light-green} backgrounds to indicate minimal performance degradation. The findings consistently show that the FroFA module positively enhances the effectiveness of our finetuning framework.
FroFA incurs negligible additional computational and memory overhead costs. As it operates directly on features of the same dimension, its computational budget remains consistent across different backbone networks.

\subsubsection{The effect of each increment step}
%
To systematically evaluate and illustrate the incremental contributions of each step within our finetuning framework, we perform experiments on multiple datasets at varying sampling ratios. The performance increment for each step is quantified and presented in Table~\ref{tab:contribution_each} as a column-wise comparison. For instance, in the CIFAR10 dataset at a 0.2\% sampling ratio, LP increases by 10.85 compared to FF. LP-FT shows a further increase of 0.49 compared to LP, while LP-LoRA exhibits an increase of 1.89 compared to LP. $\mathrm{FACT_{w.o.FroFA}}$ (without FroFA) displays an increase of 6.29 compared to LP-FT, and FACT shows a final increase of 0.27 compared to $\mathrm{FACT_{w.o.FroFA}}$. The incremental data for each step relative to the preceding step is annotated with varying shades of blue, where darker shades signify greater contributions and lighter shades indicate lesser contributions. Performance degradation is denoted by green backgrounds.
As shown in Table~\ref{tab:contribution_each}, supported by the tiered color gradients in the experimental data, the proposed multi-phase hierarchical parameter adaptation and the lightweight model for processing extracted features collectively drive substantial performance gains. In comparison, the FroFA module contributes to a lesser degree.

%
%
The comparisons from FF to LP in Table~\ref{tab:contribution_each} constitute non-incremental comparisons of different finetuning strategies, rather than progressive enhancements. Our ablation study transitions to progressive analysis beginning with LP-FT and LP-LoRA, which build upon LP. Specifically, LP-FT performs FF following LP, whereas $\mathrm{FACT_{w.o.FroFA}}$ introduces a lightweight model upon LP-FT. The complete FACT framework subsequently integrates the FroFA module as a feature augmentation component into $\mathrm{FACT_{w.o.FroFA}}$. The methodological progression and corresponding architectural modifications are illustrated in Figure~\ref{fig:THriFT}, with comprehensive implementation details provided in the preceding methodology section.

\section{Conclusion}
\label{sec:conc}

In this paper, we first define the \textit{FiAF} task and analyze the incompatibility of existing finetuning methods within active finetuning. We then introduce a three-phase hierarchical finetuning framework FACT and its variant, the $ \mathrm{L^3FACT} $. Through both theoretical analysis and empirical evaluation, we demonstrate the superiority of our finetuning framework, which result in significant performance improvements and pushing the limit of finetuning for \textit{FiAF}. Our approach has been validated across various datasets, downstream tasks, pretraining configurations, model architectures/scales, and parameter-efficient finetuning methods, comprehensively demonstrating its generalizability. Additionally, our approach improves training and parameter efficiency, featuring reduced training time and fewer trainable parameters.
\bibliographystyle{IEEEtran}
\bibliography{IEEEabrv,main}


\newpage

 

\begin{IEEEbiography}[{\includegraphics[width=1in,height=1.25in,clip,keepaspectratio]{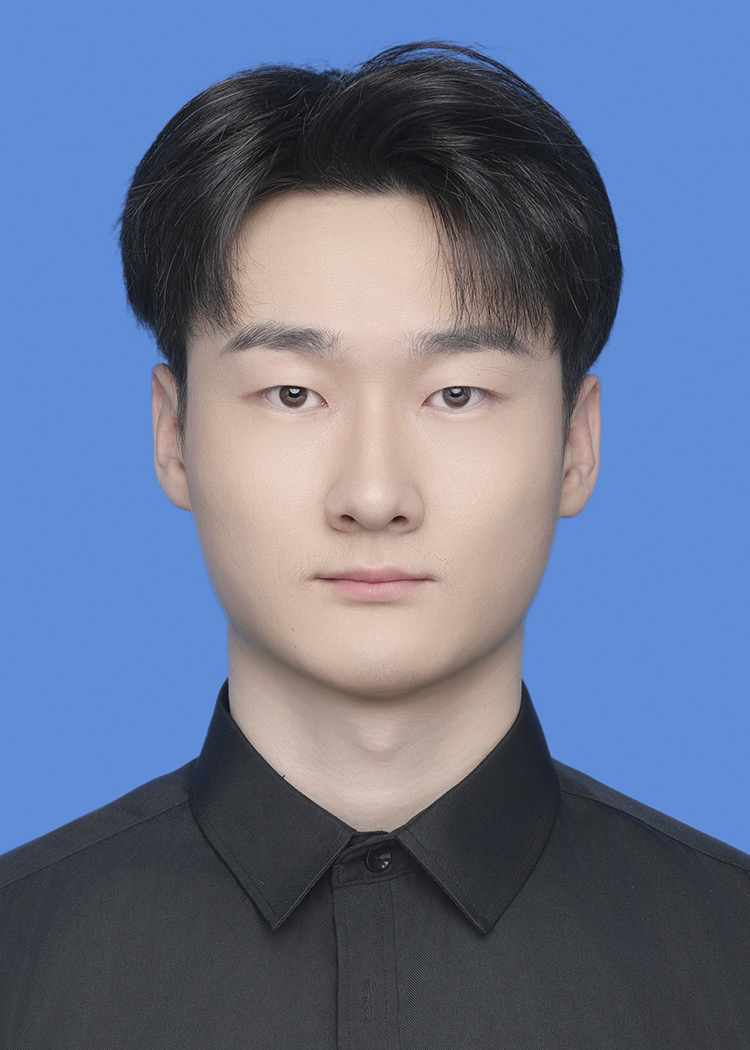}}]{Wenshuai Xu}
received the B.S. degree in software engineering from Sichuan University, Chengdu, China, in 2021 and the M.S. degree in software engineering from Beihang University, Beijing, China, in 2024. He is currently pursuing a Ph.D. degree in software engineering at Beihang University. His research interests include computer vision, image processing, and large language models.
\end{IEEEbiography}

\vspace{-0.6cm}

\begin{IEEEbiography}[{\includegraphics[width=1in,height=1.25in,clip,keepaspectratio]{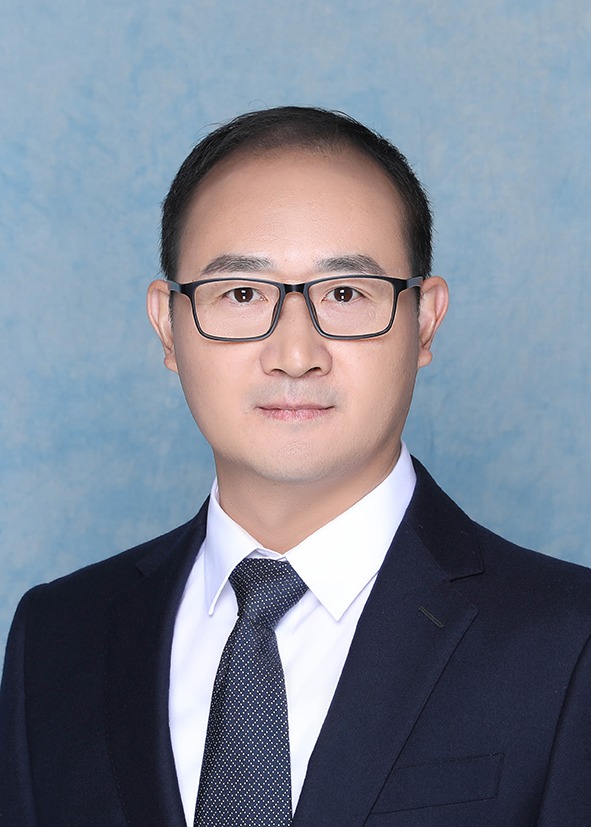}}]{You Song}
received the B.S. degree in applied mathematics and the Ph.D. degree in signal and information processing from Beihang University, in 1997 and 2003. He is a professor with the School of Software, Beihang University, China. His current research interests include data analysis techniques, information processing, knowledge graph, behavior prediction for specific fields, which are mainly applied in financial technology, online education, intelligent transportation system, bioinformatics, space science, etc.
\end{IEEEbiography}

\vspace{-0.6cm}

\begin{IEEEbiography}[{\includegraphics[width=1in,height=1.25in,clip,keepaspectratio]{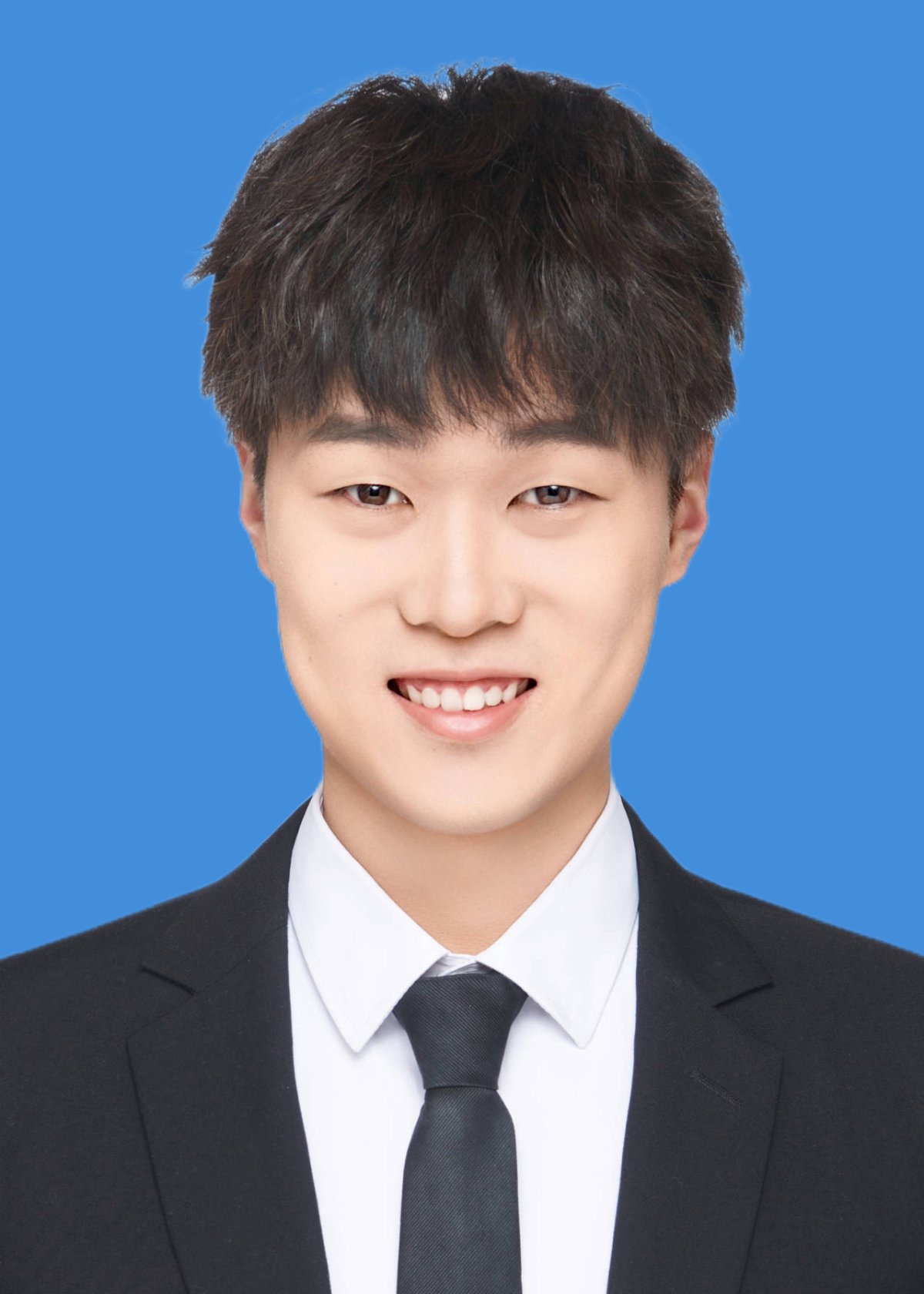}}]{Yuzhuo Cui}
received the B.S. degree in software engineering from Shandong University, Jinan, China, in 2024. He is currently a Ph.D. student at Beihang University. His research interests include graph neural networks, large language models, and efficient fine-tuning.
\end{IEEEbiography}

\vspace{-0.6cm}

\begin{IEEEbiography}[{\includegraphics[width=1in,height=1.25in,clip,keepaspectratio]{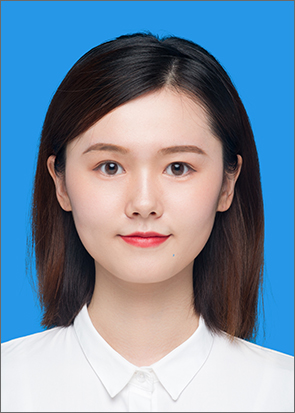}}]{Minjie Ren}
received the Ph.D. degree in electronic engineering from Tianjin University, Tianjin, China. She is currently a postdoctoral researcher at the Hangzhou Innovation Institute, Beihang University. Her research interests include multimedia signal processing, affective computing, and computer vision.
\end{IEEEbiography}

\vspace{-0.6cm}

\begin{IEEEbiography}[{\includegraphics[width=1in,height=1.25in,clip,keepaspectratio]{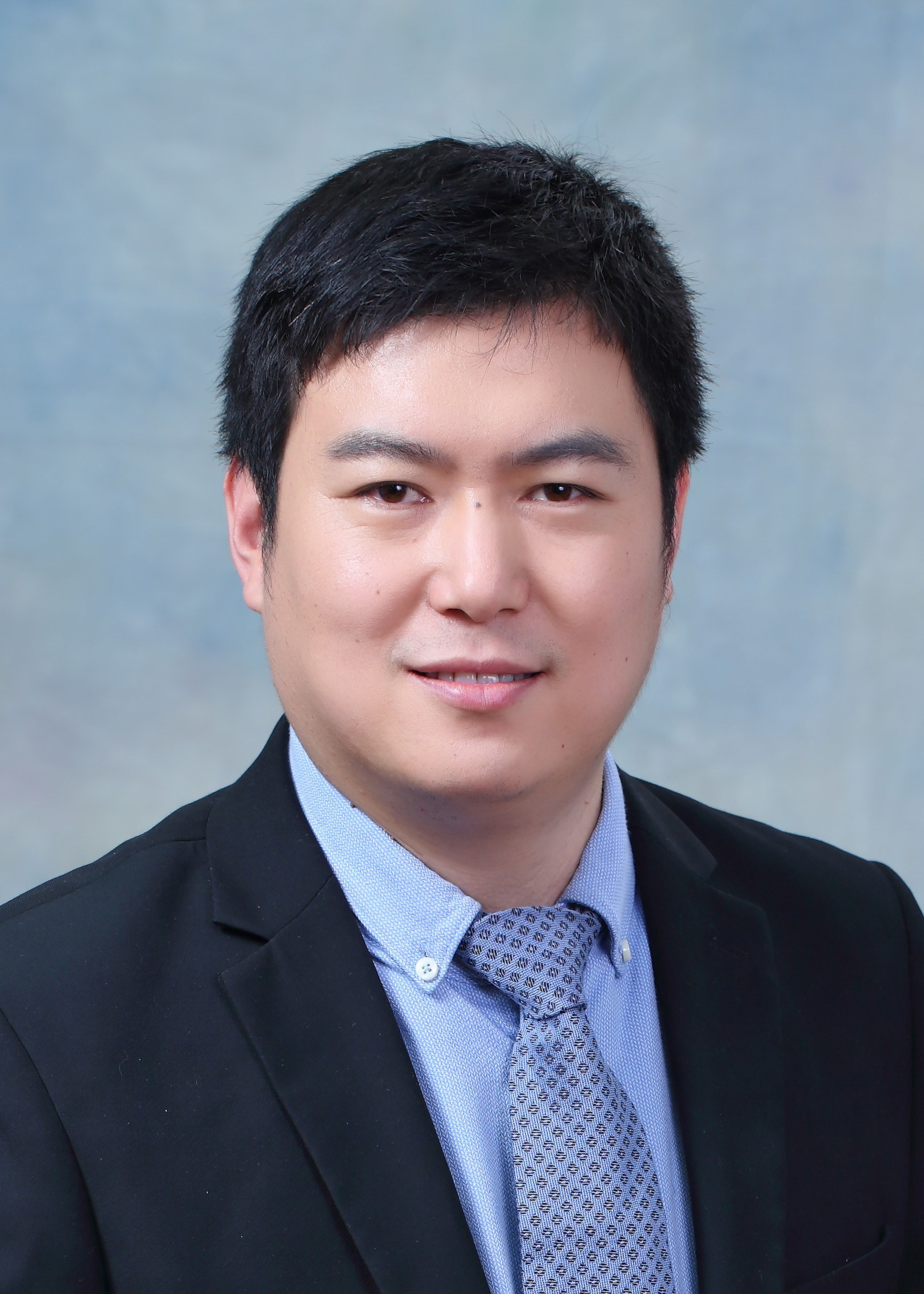}}]{Qingjie Liu}
(Member, IEEE) received the B.S. degree in computer science from Hunan University, Changsha, China, in 2007, and the Ph.D. degree in computer science from Beihang University, Beijing, China, in 2014. He is currently a professor with the School of Computer Science and Engineering, Beihang University. He is also a Distinguished Research Fellow with Hangzhou Institute of Innovation, Beihang University, Hangzhou, China. His current research interests include image fusion, object detection, image segmentation, etc.
\end{IEEEbiography}

\vspace{-0.6cm}

\begin{IEEEbiography}[{\includegraphics[width=1in,height=1.25in,clip,keepaspectratio]{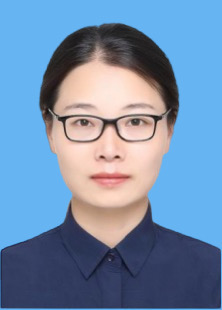}}]{Zhenghui Hu}
received the B.S. degree in computer science from the Zhejiang University of Technology, Hangzhou, China, in 2011, and the Ph.D. degree in computer science from Beihang University, Beijing, China, in 2020. She is currently a senior research associate with the Hangzhou Innovation Institute, Beihang University. Her research interests include swarm intelligence, computer vision and large language models.
\end{IEEEbiography}



\vfill

\end{document}